\documentclass[11pt,a4paper]{article}
\usepackage[hyperref]{acl2021}
\usepackage{times}
\usepackage{latexsym}

\usepackage{graphicx}
\usepackage{amsmath}
\usepackage{bm}
\usepackage{bbm}
\usepackage{amsthm}
\usepackage{amsfonts}
\usepackage{amssymb}
\usepackage{MnSymbol}

\usepackage{adjustbox}

\usepackage{microtype}

\aclfinalcopy 

\title{Relative Position Prediction as Pre-training for Text Encoders}

\author{Rickard Br\"uel-Gabrielsson \\
  MIT CSAIL \\
  \texttt{brg@mit.edu} \\\And
  Chris Scarvelis \\
  MIT CSAIL\\
  \texttt{scarv@mit.edu} \\}

\date{}

\begin{document}
\maketitle
\begin{abstract}
 Meaning is defined by the company it keeps. However, company is two-fold: It's based on the identity of tokens and also on their \textit{position} (topology). We argue that a position-centric perspective is more general and useful. The classic MLM and CLM objectives in NLP are easily phrased as position predictions over the whole vocabulary. Adapting the relative position encoding paradigm in NLP to create relative labels for self-supervised learning, we seek to show superior pretraining judged by performance on downstream tasks. 
\end{abstract}

\section{Introduction}

\emph{Word embeddings} form a core component of modern deep learning-based natural language processing. Intuitively, word embeddings map words to vectors in a semantically-meaningful manner: a desired property of word embeddings is that similarity between word representations in their embedding space (as measured by well-understood (dis)similarity measures such as cosine similarity or Euclidean distance) reflect some notion of semantic similarity. 


These embeddings are typically constructed by leveraging Harris' \emph{distributional hypothesis} \cite{harris_distributional}, which posits that words which are similar in meaning tend to often co-occur in the same contexts. This hypothesis implies that we may use essentially statistical, easily-measured properties of words as a proxy for semantic similarity, which is typically more difficult to quantify. Crucially, word co-occurrence statistics may be computed without the intervention of human annotators, enabling word embeddings to be trained in an \emph{unsupervised} manner.

Classical approaches such as latent semantic analysis (LSA) \cite{deerwester1990indexing} construct dense, low-dimensional word embedding vectors by performing a low-rank factorization (typically a singular value decomposition (SVD)) of a (possibly reweighted) term-document matrix $X \in \mathbb{R}^{|V| \times |D|}$ such that  $X_{ij}$ stores the frequency at which term $i$ occurs in document $j$.

More recently, algorithms such as word2vec \cite{word2vec} have been developed to construct word embeddings using fast and scalable algorithms such as skip-gram with negative sampling (SGNS). These embeddings are constructed by solving a classification task in a self-supervised manner: They model the probability that a word $c$ is in the \emph{context} of word $w$ as an increasing (in practice generally sigmoid) function of the dot product $\langle w,c \rangle$ of their embeddings and optimize the corresponding embeddings such that words that do often co-occur are assigned embedding vectors with large dot product.

These word embeddings learned by these algorithms possess a number of striking properties. Aside from encoding similarity in word meaning, they have been shown to encode \emph{relational} similarity: Computing the expression ($\vec{king}$) - ($\vec{man}$) + ($\vec{woman}$) results in a vector that's close to the word embedding for "queen," which matches our intuitive notion of the relationships between the words in this expression. Furthermore, there exists a body of theoretical work connecting word2vec to the factorization of a shifted pointwise mutual information matrix \cite{levy-goldberg}, and subsequent work \cite{hashimoto-metric-recovery} has framed popular word embedding algorithms as consistent methods for performing metric recovery in semantic space under an elegant random-walk based model of corpus generation.

However, word embeddings computed by word2vec and GloVe possess a number of drawbacks. Most notably, these embeddings are \emph{static}: A single embedding vector is computed for each word in the vocabulary. Given that many words in natural languages are polysemous, this is a substantial limitation, and it is reasonable to consider  word embeddings that are capable of capturing different meanings of the same word. 

This problem is partially resolved by \emph{dynamic} (contextual) word embeddings such as ELMo \cite{elmo} and BERT \cite{bert} and GPT-2 \cite{gpt2}, which learn word representations that are dependent on a word's context. This allows the word embedding model to leverage the fact that a word's meaning may be disambiguated via its context: For example, the appropriate meaning of "bank" cannot be determine from the word itself, but it \emph{can} be determined from examining its context in the sentences "I deposited the cheque at the bank" and "I went fishing by the river bank". These embeddings are trained by optimizing causal or masked language modeling (CLM/MLM) objectives: They are trained to correctly predict the next token in a sequence (CLM) or to predict a randomly-masked token (MLM). The resulting embeddings have achieved state of the art performance on a variety of downstream NLP tasks.

We propose an alternative approach based on optimizing a \emph{relative position} objective. We train a language model to predict the (signed) relative positions $p_{ij}$ of pairs of words $x_i,x_j$ in a context $x_1,...,x_k$. We believe that this approach presents several advantages over existing CLM and MLM objectives, including:

\begin{enumerate}
    \item \textit{Increased label density}. In predicting relative positions, we have $n \times n$ labels instead of $n$ (CLM) or $n \times p$ (MLM), where $n$ is the length of the sequence and $p$ is masking probability.
    \item \textit{More effective sampling}.  Examples in the current context are more likely to be relevant than other words in the vocabulary. We can improve sampling even more by (i) finding the hardest negative examples, (ii) exploiting the fact that far away positions don't have to be exact, and (iii) predicting relative positions within batches. 
    \item \textit{Stronger signal}. Many positions are distance-like and we do not need to encode our labels as one-hot targets; we can instead use continuous distance as targets, which allows us to use a variety of loss functions and encode stronger signals. (We can take inspiration from metric learning and self-supervised learning on images and 3D data). 
    \item \textit{Computational efficiency}. We avoid computing an output distribution over the complete vocabulary as in MLM and CLM.
    \item \textit{More generality}. Since relative positions have successfully been adopted in both vision and graph learning \cite{bruelgabrielsson2022rewiring}, our approach directly translates to those settings; especially where the topology is more varied than in NLP.
\end{enumerate}

\section{Related Work}
 
Self-supervised representation learning has been widely adopted in the NLP literature. Models which incorporate self-supervised learning include seminal works such as Word2Vec \cite{word2vec}, ELMo \cite{elmo}, GPT \cite{gpt}, BERT \cite{bert}, BART \cite{bart}, Electra \cite{electra}, XLNet \cite{xlnet}, and T5 \cite{t5}. These approaches rely on an objective that encourages the model to reconstruct some corrupted/noisy text of language; typically incomplete text with certain tokens missing. The primary advantage of this approach is that it enables the model to learn semantically-meaningful contextual word representations from large amounts of unlabeled text "in the wild." However, computing word probabilities in a language model typically involves computing a distribution over the entire vocabulary; for sufficiently large vocabularies (which we often encounter in practice) this is a costly operation. This problem is exacerbated in the setting of self-supervised learning for computer vision and 3D learning tasks, where the spaces of possible images or 3D configurations are infinite.

Contrastive learning \cite{nce} has been used to learn a distribution without requiring the normalizing factor over the whole vocabulary, though it requires negative examples. Other approaches \cite{siamese} use Siamese Representation Learning; this is akin to a clustering method that uses architecture asymmetries to avoid model collapse to attributing all points to a single cluster, and thus doesn't need negative examples. The approach of \cite{jigsaw} is to construct "jigsaw puzzles" from patches of an image which deep neural networks are trained to solve. Similarly, \cite{doersch2015unsupervised} learns visual representations by training a convolutional neural network to predict the relative position of randomly-chosen patches in unlabeled images. These works show that this pretext task leads to useful representations for downstream image recognition tasks. 

Our work aims to provide a useful pre-trained deep language model via a self-supervised task that essentially corresponds to reshuffling the words into the correct order. However, we do this by predicting the relative positions rather than the absolute positions; hence, achieving \textit{increased label density}, \textit{stronger signal}, and \textit{more generality}.

\section{Technical Approach}


Given a context $c=x_1,\dots,x_k$, a word $x_i, 1 \leq i \leq k$, and (relative) positions $p=p_{i,j}, 1 \leq i,j \leq k$, masked language models (MLMs) and causal language models (CLMs) attempt to correctly predict $P(x_i|c - x_i, p)$ for a training dataset, i.e. the correct word given the context and relative positions. In contrast, our position-centric approach broadly attempts to predict $P(p_{i,j}|c, p - p_{i,:} - p_{:,j})$, i.e. the correct \emph{relative positions} of pairs of words given their context and the words themselves. 

Let $F:\mathbb{R}^{k\times h} \rightarrow \mathbb{R}^{k \times h}$ be a Transformer and let $e_T: \mathcal{V} \rightarrow \mathbb{R}^d$ be the target embeddings. Then one typically models the probability of the $i$-th word given the remainder of its context as follows:

\small
\begin{equation*}
     P(x_i|c - x_i, p)=\frac{\exp(e_T(x_{i})^T F(x_1,\dots,x_k)_i)}{\sum_{x_j \in \mathcal{V}} \exp( e_T(x_j)^{T} F(x_1,\dots, x_k)_i ) } 
 \label{eq:normal_loss}
\end{equation*}

\normalsize

with appropriately masked tokens or causal masking. In our position-centric approach we model the probability of each pair of words' relative positions as follows:
\begin{multline*}
 P(p_{i,j}|c, p - p_{i,:} - p_{:,j}) \\
  = \phi(F(x_{1}, \dots, x_k)_{i}, F(x_1,\dots, x_k)_j)
 \label{eq:new_loss}
\end{multline*}

where $\phi : \mathbb{R}^{2 \times h} \rightarrow \mathcal{P}(\{p_{i,j} | i,j \in \mathbb{N} \})$ is a function to the set of all probability distributions over the relative positions. 

It is unclear a priori how we might construct $\phi$. Typically a Transformer model computes the following (for each $i$ and $j$) in its multi-head attention module:
\begin{equation*}
    \textrm{Concat}(\textrm{head}_1, \dots, \textrm{head}_{n_h})_{i,j} \in \mathbb{R}^{n_h}
\end{equation*}

where
\begin{equation*}
     \textrm{head}_l =
     (F(x_1,\dots, x_k)_i W^l_q)^T(F(x_1,\dots, x_k)_j W^l_k)
\end{equation*}

and where $W_q, W_k \in \mathbb{R}^{h \times h}$ are learnable weight matrices and $n_h$ is the number of heads.

A simple approach for constructing an appropriate function $\phi$ is to first retrieve these head-scores, corresponding to the unknown relative positions in the last attention layer of the Transformer model and append a readout layer to predict the ground truth relative positions:

\begin{equation*}
    \psi(\textrm{Concat}(\textrm{head}_1, \dots, \textrm{head}_{n_h})_{i,j})) \in \mathbb{R}^{n_p}
\end{equation*}

Here $\psi : \mathbb{R}^{n_h} \rightarrow \mathbb{R}^{n_p}$ can be a linear layer or a sequence of fully connected layers, and $n_p = |\{p_{i,j} \ | \ i,j \in \mathbb{N}\}|$. By normalizing the output of $\psi$ via a softmax function we obtain the desired probability distribution over all possible relative positions. 

\subsection{Masking or Permuting Positions}

The previous discussion adopts the perspective of masking some positions and then predicting them. An alternative approach is to \emph{permute} rather than mask positions. This raises the following questions:
\begin{enumerate}
    \item If we permute positions, should our model predict only the permuted positions, or should it also predict the non-permuted positions?
    \item If we permute positions and predict the non-permuted positions as well, should we instead consider the binary problem of predicting whether a position is correct or not?
\end{enumerate}

We call the approach of masking positions and predicting them Position-Mask-Language-Modeling (PMLM) and the approach of permuting positions are predicting the correct one for Position-Permutation-Language-Modeling (PPLM). Specifically, when permuting positions we permute a certain percentage of all tokens and we predict the correct positions of all tokens; if we predict just the permuted positions we call it PPLM-Some, and if we predict binary correct-or-not-correct positions for each position we call it PPLM-Binary.

We investigate the effectiveness of masking and these alternative permutation-based approaches empirically in the results section.

\section{Results}

During pre-training, we train on the BookCorpus dataset \cite{Zhu_2015_ICCV} and English-language Wikipedia \cite{wikidump}. We use as our transformer architecture a Roberta-Transformer \cite{roberta} with 12 layers. We employ a sequence length of 128 and batch size of 64. We train for 40K batch iterations, which takes about 4 hours on a single Tesla V100 GPU. 

We evaluate our pre-trained model's performance on three downstream tasks after fine-tuning: multi-genre natural language inference (MNLI) \cite{mnli-williams}, linguistic acceptability on the Corpus of Linguistic Acceptability (CoLA) \cite{warstadt-etal-2019-neural}, and question-answering natural language inference (QNLI) \cite{rajpurkar}. All of these tasks are components of the General Language Understanding Evaluation (GLUE) benchmark \cite{glue}. For all downstream tasks, we fine-tune our pre-trained model by following the procedure outlined in Appendix B of \cite{electra}.

The MNLI corpus is a crowd-sourced collection of 433k sentence pairs annotated with textual entailment information. Each sentence pair is assigned one of three labels: neutral, entailment, or contradiction, and the objective of the task is to correctly predict the label for each sentence pair.

The CoLA dataset is a compilation of 10.6k English word sequences annotated with a judgment of the sequence's grammatical acceptability. These sequences are drawn from works of linguistic theory. The task is to correctly predict each sentence's acceptability label.

QNLI is based on the Stanford Question Answering Dataset \cite{rajpurkar}. This dataset consists of question-paragraph pairs. Each paragraph contains a sentence which answers the associated question. QNLI modifies this dataset by converting the task to sentence-pair classification by pairing the question sentence with each sentence in the associated paragraph; the task is then to determine whether each sentence in the paragraph answers the question.

For further training details see Appendix \ref{sec:hyperparameters}. We record our results in Table \ref{table:results}.


\begin{table*}
\centering
\begin{adjustbox}{max width=\textwidth}
\begin{tabular}{llllll}
\hline
Pretraining: & MLM-30\% & PMLM-60\% & PPLM-60\% & PPLM-Some-60\% & PPLM-Binary-60\% \\
 \hline
 MNLI Accuracy: & 0.597 & 0.6123 & 0.626  & 0.590 & 0.595 \\
 CoLA Accuracy: & 0.614 & 0.618 & 0.617  & 0.618 & 0.607 \\
 QNLI Accuracy: & 0.666 & 0.642 & 0.673 & 0.631 & 0.634 \\
\hline
\end{tabular}
\end{adjustbox}
\caption{Results of Masked Language Modeling (MLM) vs versions of Positional Masked Language Modeling (PMLM) and Positional Permuted Language Modeling (PPLM).}

\label{table:results}

\end{table*}

Our results indicate that our proposed PMLM and PPLM pre-training objectives achieve competitive performance on the downstream MNLI task. The largest improvement in accuracy arises from using a model that is pre-trained using the PPLM objective with 60\% of the tokens permuted. Significant (though smaller) improvements were achieved by pre-training using the PMLM objective, which masks positions rather than permuting them.

On the other hand, pre-training using the PPLM-Some and PPLM-Binary led to small degradations in performance relative to the MLM baseline. This indicates that our model's ability to predict the actual relative position of all tokens is crucial to its ability to learn representations that lead to good performance on the downstream MNLI task. Merely predicting the relative positions of pairs of permuted words is insufficient; future work may seek to determine whether this is a byproduct of insufficient label density in the PPLM-Some problem. The PPLM is a strictly harder task than PPLM-Some and involves determining whether an ordered pair of tokens have been permuted or not, in addition to determining the ordered pairs relative position; in PPLM-Some all relative positions of the ordered pairs that the model is asked to predict have all been permuted, while this is not true for PPLM. Results show this to be a crucial difference.

Our pre-training objectives led to more muted improvements on the MLM baseline for the CoLA task. Whereas the MLM baseline achieved an accuracy of 0.614 on CoLA, our best-performing model (PMLM-60\%) achieved an accuracy of 0.618 on the same task. On the QNLI task, our best-performing model (again PPLM-60\%) achieved an accuracy of 0.673, whereas the baseline accuracy was 0.666; this is once again a notable improvement over the baseline, especially given the small scale of our experiments.

\section{Discussion}

The results that we have outlined above indicate that our relative position-based objective shows significant promise as a new method for pre-training large transformer-based language models. We conjecture that some of the benefits of our model arise from the increased label density afforded by relative position-based training: The number of labels available for our model to predict in pre-training is quadratic in the length of each context rather than linear as is the case with the standard CLM and MLM objectives. 

Further benefits arise from the increased computational efficiency of our approach. As our model does not involve the computation of an output distribution over the vocabulary but rather a distribution over the (much smaller) set of valid relative positions, our model trains more quickly using our objective and hence is able to achieve improved downstream performance given a fixed wall-clock time budget. Note that in these experiments we trained for a fixed number of iterations, not flops, and the MLM uses more flops than PPLM and PMLM. 

Finally, our model predicts relative positions of words within the same context; these words are more likely to be relevant to each other than words in the remainder of the vocabulary, potentially allowing our objective to provide a stronger signal than traditional CLM/MLM objectives that predicts words drawn from the entire vocabulary.






\section{Conclusion}

In this work, we have investigated the effectiveness of a \emph{relative position prediction} pretext task for the purpose of pre-training language models. Our results indicate that while there are some technical challenges associated with this approach, relative position prediction can lead to notable improvements in performance on downstream tasks relative to pre-training baselines based on MLM or CLM objectives. 

In future work, we hope to investigate how our model responds to increased scale. In particular, we only trained our model for 4 hours on a single Tesla V100 GPU. Modern language models are particularly likely to benefit from extended training on very large datasets \cite{gpt}, and we hope that further training will expand the performance gap between our relative position-based approach and the CLM/MLM objectives that are currently used in state-of-the-art language models. We especially hope that our objective's computational efficiency will lead to significant advantages when training models at scale; it is plausible that the small scale of our experiments in this work may have dampened some the potential benefits of our approach.

In addition, there are several alternative losses that may be used that are more distance-like; we could for example use regression losses rather than classification losses, or incorporate optimal transport distances such as Wasserstein distances between tokens.


\bibliographystyle{acl_natbib}
\bibliography{rel_positions} 

\appendix

\section{Experiment Parameters}
\label{sec:hyperparameters}

Shared hyperparameters for all experiments:

\begin{enumerate}
    \item vocab-size = 50265
    \item num-hidden-layers = 12
    \item hidden-size = 256 
    \item intermediate-size = 1024 
    \item max-position-embeddings = 128 
    \item num-attention-heads = 16 
    \item attention-probs-dropout-prob = 0.1
    \item hidden-dropout-prob = 0.1
    \item learning-rate = 5e-4
\end{enumerate}

\end{document}